\documentclass[letterpaper,10pt,conference]{ieeeconf}
\IEEEoverridecommandlockouts
\usepackage{amsmath,amssymb,booktabs,multirow}
\usepackage{graphicx}
\usepackage{xcolor}
\usepackage{url}
\usepackage{balance}
\usepackage[hidelinks]{hyperref}
\usepackage{placeins}

\title{QMSR: Query-Conditioned Mask-wise Expert Routing for Robust Open-Vocabulary Underwater Object Retrieval}

\author{Fuming Zhang*, Dongyue Huang*, Junjie Wen, Lihua Xie
\thanks{Fuming Zhang, Dongyue Huang, and Lihua Xie are with the School of Electrical and Electronic Engineering, Nanyang Technological University, Singapore 639798. {\tt\small FUMING001@ntu.edu.sg}, {\tt\small dongyue.huang@ntu.edu.sg}, {\tt\small ELHXIE@ntu.edu.sg}}
\thanks{Junjie Wen is with the Department of Mechanical and Automation Engineering, The Chinese University of Hong Kong, Shatin, N.T., Hong Kong. {\tt\small jjwen@mae.cuhk.edu.hk}}
}

\newcommand{\paperfigure}[3]{%
\begin{figure}[!t]
\centering
\IfFileExists{#1}{\includegraphics[width=\linewidth]{#1}}{%
\fbox{\parbox[c][#2][c]{0.95\linewidth}{\centering\small #3}}}
\caption{#3}
\label{fig:#1}
\end{figure}}

\begin{document}
\maketitle
\begin{abstract}
Open-vocabulary object retrieval remains challenging in complex underwater environments.
Although underwater image enhancement (UIE) can improve visual quality, fixed UIE strategies may even underperform the Raw representation in retrieval, indicating that enhancement should not be applied as a uniform preprocessing step.
To address this problem, we propose \textbf{QMSR}, a query-conditioned mask-wise expert routing framework for underwater open-vocabulary retrieval.
Specifically, QMSR selects one pretrained UIE expert for each query--candidate pair and predicts a continuous Raw--Expert fusion strength, enabling adaptive enhancement while preserving useful Raw semantics.
During training, a privileged ranking oracle provides expert-selection and fusion-strength supervision, while an annealed soft-routing relaxation facilitates optimization of the hard Top-1 routing policy.
Experiments show that QMSR improves NDCG@10 by 17.6\% over an image--query shared router, while consistently outperforming fixed UIE strategies and remaining effective on held-out query categories.
These results demonstrate the effectiveness of query-conditioned and candidate-specific enhancement routing for underwater open-vocabulary retrieval.
(Code will be released upon acceptance.)
\end{abstract}

\section{Introduction}

The complex underwater environment poses serious challenges to the visual
perception of underwater robots, limiting their capability to perform
high-level tasks such as object grasping and
manipulation~\cite{phung2024shared,wang2022underwater}. In practical
missions, the objects encountered by underwater robots may not always
belong to predefined closed-set categories~\cite{singh2025openset}.
Open-vocabulary underwater object retrieval therefore provides a flexible
way for underwater robots to identify mission-specific targets according to arbitrary
language instructions, and language-guided perception has begun to be
explored in marine robotic systems~\cite{thengane2025merlion}. Related
multimodal embodied perception has also been explored in robotic
systems~\cite{qi2026airembodied}. With
vision--language models such as CLIP~\cite{radford2021clip} and
class-agnostic proposal models such as SAM~\cite{kirillov2023sam}, target
objects can be specified at deployment time without retraining a
task-specific closed-set detector. Recent methods such as
ConceptFusion~\cite{jatavallabhula2023conceptfusion} and
PLAF~\cite{wen2026plaf} further support language-aligned region-level
perception. However, underwater attenuation, scattering, color distortion,
and low contrast can distort these representations and degrade semantic
matching between candidate regions and language queries.

\begin{figure}[!t]
\centering
\includegraphics[width=\linewidth]{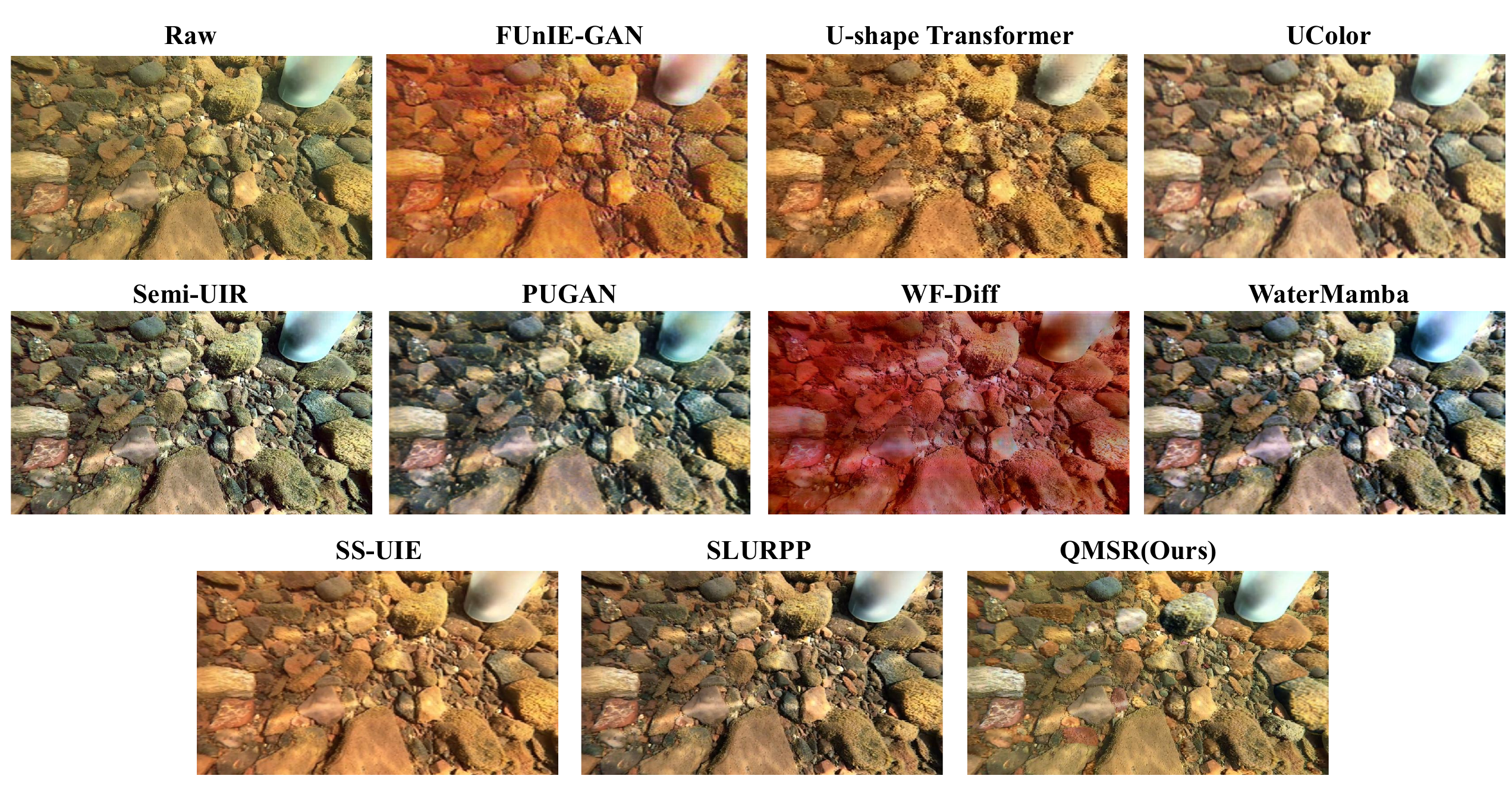}
\caption{Visual comparison of Raw, nine UIE experts, and QMSR for the
language query \emph{Plastic Cup}. Different enhancement methods produce
substantially different restoration behaviors for the same underwater
scene.}
\label{fig:expert_comparison}
\end{figure}

Underwater image enhancement (UIE) provides a natural way to alleviate
such degradation and has achieved substantial progress~\cite{akkaynak2019seathru,islam2020funie,li2021ucolor,
peng2023ushape,zhao2024wfdiff,guan2024watermamba}. Nevertheless,
better visual restoration does not necessarily imply better
open-vocabulary retrieval. Different UIE methods modify color, contrast,
texture, and structural cues differently, producing different shifts in
the vision--language feature space. In our experiments, all nine fixed UIE
experts underperform the Raw representation despite producing visually
enhanced results. Enhancement therefore should not be treated as
universally beneficial, but should depend on whether the induced feature
change benefits the current retrieval objective. Fig.~\ref{fig:expert_comparison}
illustrates the diverse behaviors of different UIE experts.

\begin{figure}[!t]
\centering
\includegraphics[width=0.80\linewidth]{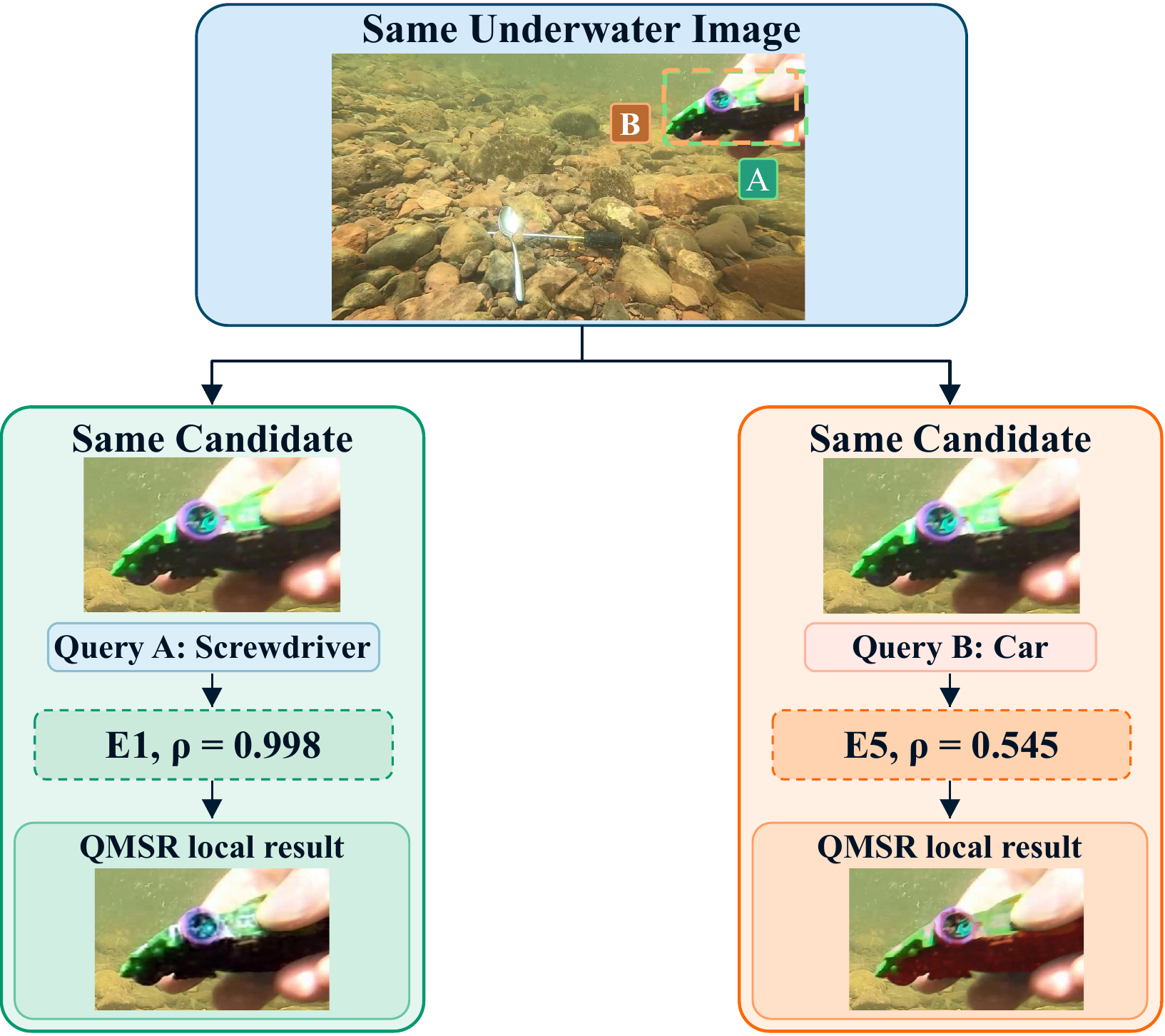}
\caption{Query-dependent enhancement behavior on the same underwater image. Different language queries favor different enhancement experts and fusion strengths, resulting in different retrieval outcomes while the underlying image remains unchanged. E1 denotes U-Shape Transformer ($\rho = 0.998$), and E5 denotes WF-Diff ($\rho = 0.545$).
}
\label{fig:different_queries}
\end{figure}

This mismatch becomes more pronounced in open-vocabulary retrieval because
the desired enhancement is not determined by the image alone. First,
different language queries may rely on different semantic cues, making the
same enhancement useful for one query but ineffective or harmful for
another. Existing task-aware enhancement methods~\cite{lin2026dtiuie,fan2026semantic} introduce task-related supervision,
while recent underwater open-vocabulary approaches~\cite{li2026maris,li2026earth2ocean} exploit vision--language
representations for flexible semantic perception. However, task-aware
enhancement is typically optimized for a predefined objective, whereas
open-vocabulary retrieval must adapt to arbitrary deployment-time queries.
Fig.~\ref{fig:different_queries} shows that different queries on the same
image can favor different enhancement decisions.

Second, query dependence alone is insufficient. Underwater degradation is
spatially non-uniform, and different candidates in the same image may
preserve different amounts of query-relevant information. Under the same
query, one candidate may benefit from enhancement, while another may
already have a reliable Raw representation or be harmed by the same
operation. Most UIE pipelines provide a shared enhanced image to all
candidate regions, while image--query-level routing similarly applies the
same decision to all masks. Neither captures candidate-specific
enhancement utility. The relevant decision is therefore which enhancement
is useful for each \emph{query--candidate pair}.

A further difficulty is that enhancement should not be forced once an
expert is selected. The Raw feature may already preserve reliable semantic
information, while fully replacing it with an enhanced representation may
introduce semantic drift. This motivates using pretrained UIE models as
heterogeneous enhancement behaviors and learning both \emph{which}
behavior is useful and \emph{how strongly} the Raw representation should
be modified. Raw semantics can thus be preserved when enhancement is
unnecessary or harmful.

Based on these observations, we propose \textbf{QMSR}, a
Query-Conditioned Mask-wise Expert Routing framework for robust
open-vocabulary underwater object retrieval. QMSR performs routing at the
semantic candidate level, assigning each SAM-generated mask an enhancement
decision conditioned on its Raw feature, the language query, and
candidate-set context. A hard single-expert router selects one pretrained
UIE expert for each candidate, while a continuous Raw--Expert strength
controls its deviation from the Raw representation. During training, a
privileged ranking oracle provides expert-preference supervision using
enhanced features and relevance information, while an annealed soft-routing
relaxation facilitates optimization of the discrete Top-1 routing policy.
At inference, the router uses only Raw candidate features, the language
feature, and candidate-set context.

Our main contributions are as follows:
\begin{itemize}
    \item We formulate underwater enhancement for open-vocabulary
    retrieval as a query-conditioned, candidate-specific routing problem,
    allowing different candidates under the same image and query to receive
    different enhancement decisions according to their retrieval utility.

    \item We introduce a Raw-anchored hard single-expert routing mechanism
    over heterogeneous pretrained UIE models, which selects one expert for
    each query--candidate pair and continuously controls its contribution
    relative to the Raw representation.

    \item We develop a privileged ranking oracle and an annealed
    soft-routing relaxation to provide ranking-oriented supervision for
    optimizing the discrete Top-1 routing policy without privileged
    information at inference. Experiments further demonstrate that QMSR
    outperforms fixed enhancement strategies and image--query shared
    routing, while controlled analyses verify the necessity of
    query-dependent and mask-specific routing decisions.
\end{itemize}

\FloatBarrier
\section{Related Work}

\subsection{Mixture-of-Experts and Conditional Routing}

Mixture-of-Experts (MoE) enables different inputs to activate different
experts through learned routing. Early sparse MoE models~\cite{shazeer2017moe} introduced conditional expert activation, while
Switch Transformer~\cite{fedus2022switch} simplified routing with a Top-1
policy. This idea was later extended to visual and multimodal learning.
V-MoE~\cite{riquelme2021vmoe} routes visual tokens to specialized experts,
and LIMoE~\cite{mustafa2022limoe} applies sparse routing to joint
language--image learning. More recently, RouterRetriever~\cite{lee2025routerretriever} performs query-level expert selection for
retrieval, while Argus-Retriever~\cite{abdallah2026argus} further
introduces query-conditioned region-aware routing.

These methods show the effectiveness of input-dependent routing, but mainly route learned representation experts over queries, tokens, or predefined regions rather than ranked semantic candidates. They therefore do not address candidate-specific enhancement needs under the same query or whether a given enhancement benefits each candidate's semantic representation.

\subsection{Adaptive and Region-Aware Image Restoration}

Image restoration has increasingly adopted adaptive mechanisms to handle
diverse and spatially varying degradations. AirNet~\cite{li2022airnet}
learns degradation-aware representations, while PromptIR~\cite{potlapalli2023promptir} uses degradation-dependent prompts.
Spatial expert mechanisms address heterogeneous distortions within an
image~\cite{kim2020spatialmoe}, and M2Restore~\cite{wang2025m2restore} explores mixture-of-experts for all-in-one image
restoration. Similar ideas have been introduced into underwater
restoration. UniUIR~\cite{zhang2025uniuir} models multiple underwater
degradations with a mixture of experts, while CoRe-UIE~\cite{kong2026coreuie} performs region-adaptive routing for coexisting
underwater degradations.

These approaches model region-specific restoration, but routing is mainly degradation-driven and optimized for perceptual quality rather than language-guided retrieval. Enhancement may improve visual quality while distorting query-relevant color, texture, or structural cues, and its effect can vary across queries, with Raw sometimes remaining preferable. Thus, restoration-oriented routing does not explicitly model query-dependent enhancement utility.

\subsection{Visual Enhancement via Task Semantics and Text Prompts}

Task-aware enhancement further considers whether restoration benefits
downstream perception. IA-YOLO~\cite{liu2022iayolo} learns image-adaptive
processing jointly with object detection, while EnYOLO~\cite{wen2024enyolo} integrates underwater enhancement and detection.
Recent underwater methods~\cite{lin2026dtiuie,fan2026semantic} incorporate
semantic supervision, and AMIEOD~\cite{huang2026amieod} introduces
multiple enhancement experts with detection-oriented selection. Language
has also been used to guide restoration. InstructIR~\cite{conde2024instructir} uses natural-language instructions to specify
restoration behavior, while text-guided methods~\cite{liu2025textguided}
explore semantic alignment for cross-modal applications. In parallel,
CLIP~\cite{radford2021clip}, ConceptFusion~\cite{jatavallabhula2023conceptfusion}, and recent underwater
open-vocabulary methods~\cite{wen2026plaf,li2026maris,li2026earth2ocean}
provide language-aligned representations for open-vocabulary perception.

However, task-aware enhancement is typically optimized for a fixed downstream objective such as detection, while instruction-guided restoration uses language mainly to specify restoration behavior. Open-vocabulary retrieval instead requires arbitrary deployment-time queries to determine candidate rankings. Enhancement decisions must therefore depend jointly on the query and candidate, with utility measured by retrieval ranking rather than restoration quality alone.

Existing methods do not directly address this setting. QMSR formulates enhancement selection as query-conditioned, candidate-specific routing over heterogeneous pretrained UIE models. The fixed expert pool preserves distinct restoration behaviors, while the router selects useful enhancement for each query--candidate pair. SAM masks serve as both object-aligned routing units and retrieval candidates, and a continuous Raw--Expert strength preserves Raw features when enhancement is unnecessary or harmful. The routing policy is optimized according to candidate ranking, directly aligning enhancement decisions with open-vocabulary retrieval rather than perceptual quality or a fixed downstream task.

\section{Methodology}

\begin{figure*}[!t]
\centering
\includegraphics[width=\textwidth]{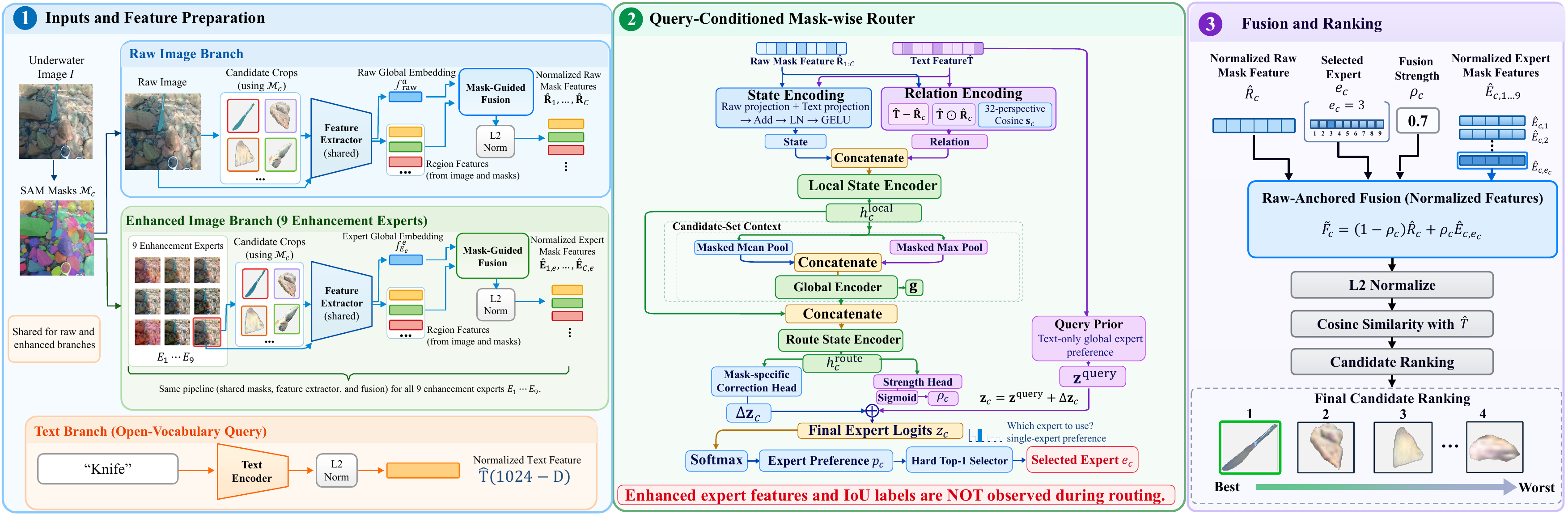}
\caption{Overall framework of QMSR. The deployment path performs
query-conditioned mask-wise routing followed by Raw-anchored fusion and
candidate ranking. The privileged Oracle and soft-routing branch are used
only during training.}
\label{fig:qmsr_framework}
\end{figure*}

The overall framework of QMSR is shown in Fig.~\ref{fig:qmsr_framework}.
Given an underwater image and a language query, SAM generates candidate
masks and a frozen vision--language backbone extracts language-aligned
features. A query-conditioned mask-wise router predicts one enhancement
expert and a continuous Raw--Expert strength for each candidate. The
selected expert feature is then fused with the Raw feature and ranked
against the text feature. During training, a privileged ranking Oracle and
an annealed soft-routing branch are introduced to facilitate hard routing
optimization.

\subsection{Problem Formulation}

Given an underwater image $I$, SAM~\cite{kirillov2023sam} produces $C$
candidate masks. For candidate $c$, the frozen vision--language backbone
extracts a Raw feature $\mathbf R_c\in\mathbb R^{1024}$ and nine enhanced
features $\mathbf E_{c,e}\in\mathbb R^{1024}$, $e=1,\ldots,9$. The
language query is encoded as $\mathbf T\in\mathbb R^{1024}$.

During training, $Y_c$ denotes the IoU relevance of candidate $c$. The
deployed router observes only Raw candidate features, the text feature,
and candidate-set context. Ground-truth IoU is used exclusively for training.
Enhanced features are excluded from the router inputs, but the selected
expert feature is required for inference-time fusion and retrieval scoring.
IoU defines relevance supervision, while the final retrieval score is always
computed from the routed visual feature and the text feature.

\subsection{Query-Conditioned Mask-wise Router}

The router models both query--candidate relations and candidate-set
context. Raw and text features are first $\ell_2$ normalized. Their
relation is encoded using feature difference, element-wise product, and
learned multi-perspective cosine similarity:
\begin{equation}
\mathbf h_c^{\mathrm{rel}}
=
f_{\mathrm{rel}}
\left(
[
\hat{\mathbf T}-\hat{\mathbf R}_c;
\hat{\mathbf T}\odot\hat{\mathbf R}_c;
\mathbf s_c
]
\right),
\end{equation}
where
\begin{equation}
s_{c,k}
=
\cos
\left(
\mathbf w_k\odot\hat{\mathbf T},
\mathbf w_k\odot\hat{\mathbf R}_c
\right).
\end{equation}

The Raw--text state and relation representation are encoded into a local
feature $\mathbf h_c^{\mathrm{local}}$. Masked mean and max pooling over valid
candidates construct a permutation-invariant global context $\mathbf g$,
which is combined with the local feature as
\begin{equation}
\mathbf h_c^{\mathrm{route}}
=
f_{\mathrm{route}}
\left(
[
\mathbf h_c^{\mathrm{local}};
\mathbf g
]
\right).
\end{equation}

To separate query-level preference from candidate-specific adjustment,
the routing logits are decomposed as
\begin{equation}
\mathbf z_c
=
\mathbf z^{\mathrm{query}}
+
\Delta\mathbf z_c,
\qquad
\mathbf p_c=\operatorname{Softmax}(\mathbf z_c),
\end{equation}
where $\mathbf z^{\mathrm{query}}$ depends only on the text feature and
$\Delta\mathbf z_c$ is predicted from $\mathbf h_c^{\mathrm{route}}$.

The router also predicts an individual enhancement strength
\begin{equation}
\rho_c
=
\operatorname{sigmoid}
\left(
f_\rho(\mathbf h_c^{\mathrm{route}})
\right)
\in[0,1].
\end{equation}
Thus, each query--candidate pair obtains both an expert preference and an
enhancement strength.

\subsection{Hard Single-Expert Raw-Anchored Routing}

At inference, QMSR selects exactly one expert:
\begin{equation}
e_c=\arg\max_e p_{c,e}.
\end{equation}

At inference, if $\rho_c<\tau_\rho$, we set $\rho_c=0$ and directly
retain the Raw representation.

The selected expert feature is fused with the Raw feature as
\begin{equation}
\widetilde{\mathbf F}_c
=
(1-\rho_c)\hat{\mathbf R}_c
+
\rho_c\hat{\mathbf E}_{c,e_c}.
\end{equation}
The retrieval score is
\begin{equation}
S_c
=
\cos
\left(
\widetilde{\mathbf F}_c,
\hat{\mathbf T}
\right),
\end{equation}
and candidates are ranked according to $S_c$.

The Raw anchor allows the router to continuously control the influence of
enhancement, so each deployed routing action consists only of one expert
identity and one fusion strength.

\subsection{Privileged Ranking Oracle}

Direct supervision for expert selection is unavailable from retrieval
annotations. We therefore introduce a training-only privileged Oracle,
implemented as offline optimization rather than an additional neural
network. For each training image--query sample, it observes the Raw
features, all nine expert features, the text feature, and ground-truth IoU
relevance.

For candidate $c$, the Oracle optimizes an unrestricted Raw+$9$ expert
composition:
\begin{equation}
\widetilde{\mathbf F}_c^{\mathrm{oracle}}
=
\left(1-\sum_e W_{c,e}\right)\hat{\mathbf R}_c
+
\sum_e W_{c,e}\hat{\mathbf E}_{c,e},
\end{equation}
where $W_{c,e}\geq 0$ and $\sum_e W_{c,e}\leq 1$.

Given the composition weights $W_{c,e}$, the Oracle computes the
retrieval score
\begin{equation}
S_c^{\mathrm{oracle}}
=
\cos
\left(
\widetilde{\mathbf F}_c^{\mathrm{oracle}},
\hat{\mathbf T}
\right).
\end{equation}
The weights $\{W_{c,e}\}$ of all candidates are then jointly optimized
by the LambdaNDCG@10 objective using IoU $Y_c$ as graded relevance.
The optimized weights $W_{c,e}^*$ therefore define the privileged
enhancement action for each candidate.

The Oracle target is decomposed into enhancement strength and expert
preference:
\begin{equation}
\rho_c^*
=
\sum_eW_{c,e}^*,
\qquad
q_{c,e}^*
=
\frac{W_{c,e}^*}{\rho_c^*}.
\end{equation}
where $\mathbf q_c^*$ is defined for the active set
$\mathcal{A}=\{c\mid\rho_c^*>0\}$.
These targets supervise the student router during training. The Oracle is
removed entirely at inference, where routing must be inferred only from
Raw and text information.

\FloatBarrier

\begin{figure*}[!t]
\centering
\includegraphics[width=0.94\textwidth]{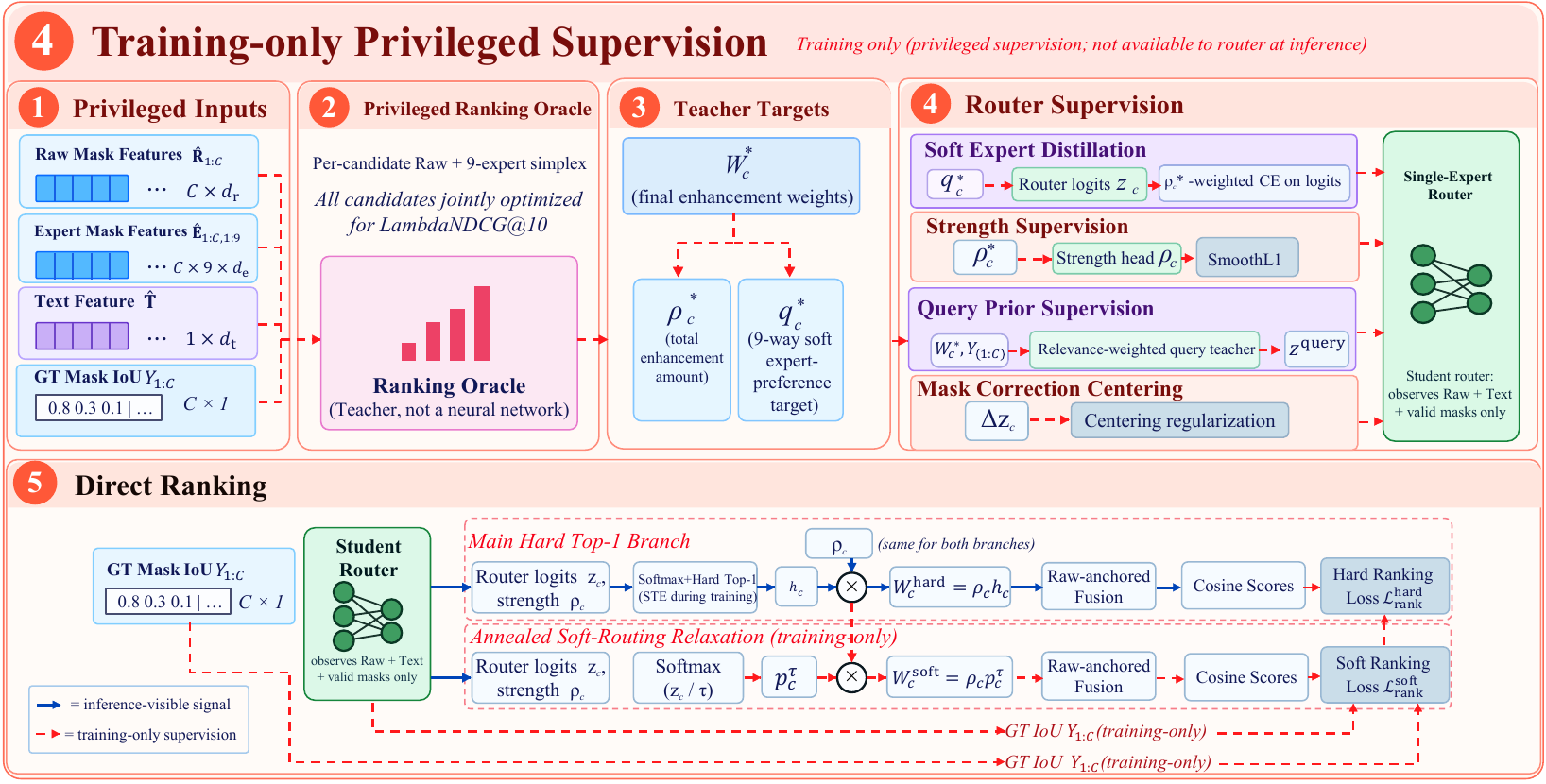}
\caption{Training strategy of QMSR. The privileged Oracle and annealed soft-ranking branch provide training-only supervision for the hard Top-1 router and are removed at inference. Enhanced features are excluded from router inputs, while the selected expert feature is retained for inference-time fusion and retrieval scoring.}
\label{fig:qmsr_training}
\end{figure*}

\subsection{Ranking Optimization}

Since hard Top-1 selection is non-differentiable, a straight-through
estimator (STE) is used during training:
\begin{equation}
\widetilde{\mathbf h}_c
=
\mathbf h_c
+
\mathbf p_c
-
\operatorname{sg}(\mathbf p_c),
\end{equation}
where $\mathbf h_c$ is the one-hot Top-1 vector and sg denotes the
stop-gradient operator. The forward path remains hard, while gradients
are propagated through $\mathbf p_c$.

An auxiliary soft-routing branch is used during the early training stage:
\begin{equation}
\mathbf p_c^\tau
=
\operatorname{Softmax}
\left(
\frac{\mathbf z_c}{\tau}
\right),
\end{equation}
which replaces the hard expert choice with a weighted expert combination
and serves only as a training relaxation.

The hard branch uses the STE in Eq.~(12) for Top-1 routing and Eq.~(7)
for single-expert fusion, while the soft branch uses the
$\mathbf p_c^\tau$-weighted expert combination in Eq.~(13); both compute
scores as in Eq.~(8).

The Oracle supervises expert preference and enhancement strength through
\begin{equation}
\begin{aligned}
\mathcal{L}_{\mathrm{distill}}
&=
-\frac{
\sum_{c\in \mathcal{A}}
\rho_c^*
\sum_e q_{c,e}^*\log p_{c,e}
}{
\sum_{c\in \mathcal{A}}\rho_c^*
}, \\[3pt]
\mathcal{L}_{\rho}
&=
\operatorname{SmoothL1}(\rho_c,\rho_c^*).
\end{aligned}
\end{equation}

Both hard and soft routing branches are optimized by LambdaNDCG@10:
\begin{equation}
\mathcal{L}_{\mathrm{rank}}
=
\sum_{Y_i>Y_j}
|\Delta\mathrm{NDCG}_{ij}|
\operatorname{softplus}
[-\gamma(S_i-S_j)],
\end{equation}
where $\gamma$ is the pairwise score scaling factor.

Although IoU defines the training relevance, QMSR does not learn an
independent ranking predictor. The score $S_c$ is determined only by
expert selection, fusion strength, and visual--text similarity. The
ranking loss therefore optimizes the routing action rather than directly
regressing retrieval scores from IoU.

We additionally use $\mathcal{L}_{\mathrm{prior}}$ to supervise the query
prior with relevance-weighted Oracle preferences and
$\mathcal{L}_{\mathrm{center}}$ to center the relevance-weighted
mask-specific corrections.

The overall objective is
\begin{equation}
\begin{aligned}
\mathcal{L}
={}&
\lambda_{\mathrm{h}}\mathcal{L}_{\mathrm{rank}}^{\mathrm{hard}}
+\lambda_{\mathrm{a}}\mathcal{L}_{\mathrm{rank}}^{\mathrm{soft}}
+\lambda_{\mathrm{d}}\mathcal{L}_{\mathrm{distill}}
+\lambda_\rho \mathcal{L}_\rho \\
&+
\lambda_{\mathrm{p}}\mathcal{L}_{\mathrm{prior}}
+\lambda_{\mathrm{c}}\mathcal{L}_{\mathrm{center}}.
\end{aligned}
\end{equation}

The temperature and soft-routing contribution are gradually annealed
during training, and the soft branch is removed in the final stage.
Consequently, inference uses only hard Top-1 routing and Raw-anchored
fusion.

\section{Experiments}

In this section, we first introduce the experimental settings
(Sec.~\ref{sec:implementation}). Then, the proposed QMSR is compared with fixed enhancement
strategies and an image--query shared router (Sec.~\ref{sec:quantitative}).
Its generalization to category-disjoint unseen queries is further evaluated in Sec.~\ref{sec:unseen}.
The effectiveness
of query-conditioned and mask-wise routing is further investigated by
controlled routing-transfer experiments (Sec.~\ref{sec:routing_effectiveness}). Finally, the routing
action space and the training strategy are analyzed in Sec.~\ref{sec:action_space} and
Sec.~\ref{sec:ablation}, respectively.

\subsection{Implementation Details}
\label{sec:implementation}

We conduct experiments on COU~\cite{mukherjee2025cou},
TrashCan~\cite{hong2020trashcan}, UIIS10K~\cite{li2025uiis10k}, and
USIS10K~\cite{lian2024usis}. The processed cache contains 39,390,
6,835, and 6,377 image--query samples for training, validation, and test,
respectively. The test split contains 1,382 COU, 3,115 UIIS10K, and
1,880 USIS10K samples. Each sample retains its complete candidate set
without truncation, with an average of 84.18 candidates in the test set.

The nine enhancement experts are FUnIE-GAN~\cite{islam2020funie},
U-Shape Transformer~\cite{peng2023ushape}, UColor~\cite{li2021ucolor},
Semi-UIR~\cite{huang2023semiuir}, PUGAN~\cite{cong2023pugan},
WF-Diff~\cite{zhao2024wfdiff}, WaterMamba~\cite{guan2024watermamba},
SS-UIE~\cite{peng2025ssuie}, and SLURPP~\cite{wu2025slurpp}.
The feature and router hidden dimensions are 1024 and 256, respectively.
The model is trained with AdamW using a batch size of 16 and an initial
learning rate of $2\times10^{-4}$, with 5\% linear warmup followed by
cosine decay. Training is stopped according to validation NDCG@10, with
the final model stopping at epoch 86.

Since open-vocabulary retrieval produces a ranked candidate list,
NDCG@10 is adopted as the primary metric. Candidate IoU is used as graded
relevance, so NDCG rewards high-overlap masks at higher ranks and
normalizes the gain by the ideal ranking. We additionally report NDCG@5,
Top-1 IoU, and Recall@1/3 at an IoU threshold of 0.5. The Raw fallback
threshold is calibrated only on the validation split and fixed to
$\tau_\rho=0.23$ for final test evaluation.

\subsection{Quantitative Comparison}
\label{sec:quantitative}

To examine whether underwater enhancement itself improves
open-vocabulary retrieval, we first compare Raw features with fixed
enhancement strategies. Among the nine evaluated fixed UIE experts, SLURPP achieves the highest NDCG@10 of 0.4578, below the 0.4655 obtained by Raw (Table~\ref{tab:main_results}). Uniformly
combining all experts reaches 0.4666, only 0.24\% above Raw. These results
show that improved perceptual quality does not necessarily translate into
better open-vocabulary retrieval.

\begin{table}[!t]
\centering
\caption{Quantitative comparison on the test split.}
\label{tab:main_results}
\small
\setlength{\tabcolsep}{3.4pt}
\begin{tabular}{lccccc}
\toprule
Method & N@5 & N@10 & Top1 & R@1 & R@3 \\
\midrule
Raw & 0.4275 & 0.4655 & 0.3446 & 0.4058 & 0.5701 \\
Best Fixed (SLURPP) & 0.4215 & 0.4578 & 0.3393 & 0.3982 & 0.5675 \\
Uniform-9 & 0.4296 & 0.4666 & 0.3517 & 0.4151 & 0.5712 \\
Image--Query Shared & 0.4570 & 0.4920 & 0.3749 & 0.4431 & 0.6070 \\
\textbf{QMSR} & \textbf{0.5454} & \textbf{0.5786} &
\textbf{0.4397} & \textbf{0.5207} & \textbf{0.6948} \\
\bottomrule
\end{tabular}

\vspace{2pt}
\parbox{\linewidth}{\footnotesize
\emph{Note:} N@5/N@10 denote NDCG@5/NDCG@10, and
R@1/R@3 denote Recall@1/Recall@3.}
\end{table}

\begin{figure}[!t]
\centering
\includegraphics[width=\linewidth]{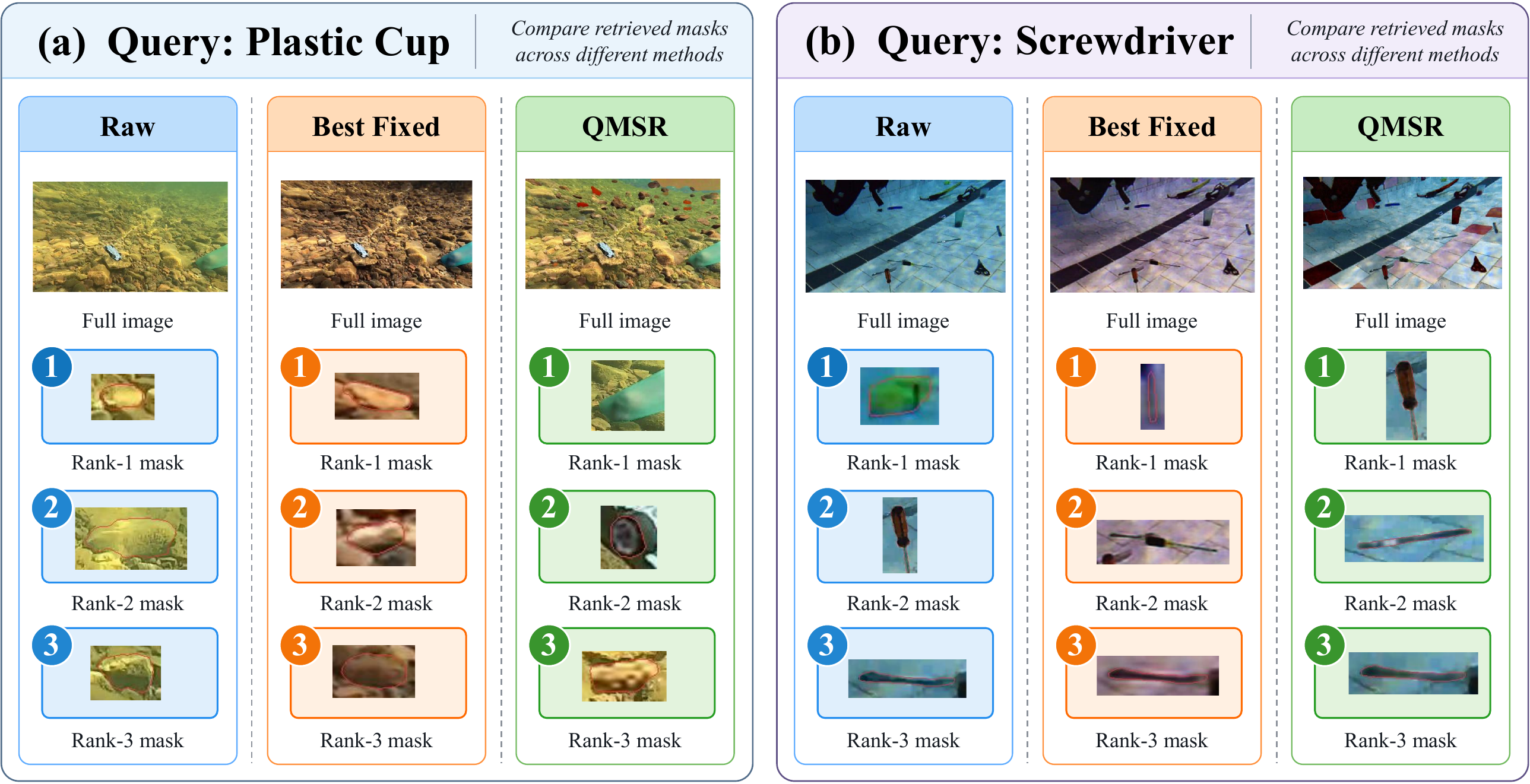}
\caption{Representative retrieval results under different language queries
and candidate masks. Different query--mask pairs favor different
enhancement behaviors, while QMSR adapts the routing decision to each
candidate.}
\label{fig:query_mask_examples}
\end{figure}

We further train an Image--Query Shared router that predicts a single
routing decision for all candidate masks within the same image--query
pair. It improves NDCG@10 to 0.4920, but remains substantially below the
0.5786 achieved by QMSR. Compared with the best fixed UIE expert and the
shared router, QMSR improves NDCG@10 by 26.4\% and 17.6\%, respectively.
These results support adapting enhancement decisions to individual query–candidate pairs. Representative retrieval results are shown in
Fig.~\ref{fig:query_mask_examples}.

\FloatBarrier

\subsection{Generalization to Unseen Queries}
\label{sec:unseen}

To further evaluate open-vocabulary generalization, we construct a
category-disjoint split in which test query categories are excluded from
router training and validation. Among 59 query labels, 51 are used as seen
categories and eight are held out: \emph{arthropoda}, \emph{Lure},
\emph{Plastic Cup}, \emph{robots}, \emph{ruins}, \emph{Scissors},
\emph{Soda Can}, and \emph{Spoon}. The split contains 34,571 seen training
samples, 5,768 seen validation samples, 5,603 seen test samples, and 774
unseen test samples. The eight held-out query labels do not appear in router training or validation.

The same router architecture and training strategy are used without any
unseen-query samples. Training stops at epoch 71 with a best seen-validation
NDCG@10 of 0.549. The Raw-fallback threshold is calibrated only on the
seen validation set and fixed at $\tau_\rho=0.14$ for both seen and unseen
evaluation. SLURPP, the best fixed UIE expert in
Table~\ref{tab:main_results}, is used as the fixed-enhancement baseline.

\begin{table}[!t]
\centering
\caption{Category-disjoint evaluation on seen and unseen queries.}
\label{tab:unseen_results}
\small
\setlength{\tabcolsep}{4pt}
\begin{tabular}{lcc}
\toprule
Method & \shortstack{Seen\\NDCG@10} & \shortstack{Unseen\\NDCG@10} \\
\midrule
Raw & 0.461947 & 0.491151 \\
Best Fixed (SLURPP) & 0.456446 & 0.467713 \\
\textbf{QMSR} & \textbf{0.575412} & \textbf{0.509722} \\
\bottomrule
\end{tabular}
\end{table}
On seen queries, QMSR achieves 0.5754 NDCG@10, improving by 26.06\%
over the best fixed UIE expert at 0.4564. On the eight held-out
categories, QMSR achieves 0.5097 compared with 0.4677 for the same
baseline, corresponding to an 8.98\% improvement. These results show that
query-conditioned mask-wise routing remains effective for query categories
completely excluded from router training and validation.

\subsection{Effectiveness of Query-Conditioned Mask-wise Routing}
\label{sec:routing_effectiveness}

We further examine whether routing decisions depend on both the candidate
mask and the language query. Since the final Top-1 router does not expose a
continuous expert composition, we use the unrestricted analysis router to
separately control expert preference $\mathbf q$ and enhancement strength
$\rho$.

For cross-mask transfer, the target Raw and expert features are fixed,
while routing parameters are transferred from another mask in the same
image--query pair. As shown in Table~\ref{tab:routing_transfer}, replacing
the expert preference increases the target rank by 2.120 on average,
compared with 0.871 when transferring only the strength, making the former
$2.43\times$ more harmful.

For cross-query transfer, the image and candidate masks remain unchanged,
while routing parameters from another valid query are applied. Across
4,592 transfer events from 1,498 image groups, replacing $\mathbf q$
increases the rank by 3.210, compared with 0.863 for $\rho$, making the
expert-preference mismatch $3.72\times$ more harmful. Transferring both
parameters from the wrong query increases the rank by 6.557 on average.
Image-grouped bootstrap confidence intervals are strictly positive for all
interventions.

\begin{table}[!t]
\centering
\caption{Controlled routing-transfer analysis.}
\label{tab:routing_transfer}
\small
\setlength{\tabcolsep}{4.0pt}
\begin{tabular}{llcc}
\toprule
Setting & Transferred Part & $\Delta$Rank & Top-10 Loss \\
\midrule
\multirow{3}{*}{Cross-mask}
 & $\rho$ & 0.871 & 2.63\% \\
 & $\mathbf q$ & 2.120 & 6.72\% \\
 & $\rho+\mathbf q$ & 4.136 & 11.38\% \\
\midrule
\multirow{3}{*}{Cross-query}
 & $\rho$ & 0.863 & 2.94\% \\
 & $\mathbf q$ & 3.210 & 8.01\% \\
 & $\rho+\mathbf q$ & 6.557 & 15.07\% \\
\bottomrule
\end{tabular}
\end{table}

\begin{figure*}[!t]
\centering
\includegraphics[width=0.84\textwidth]{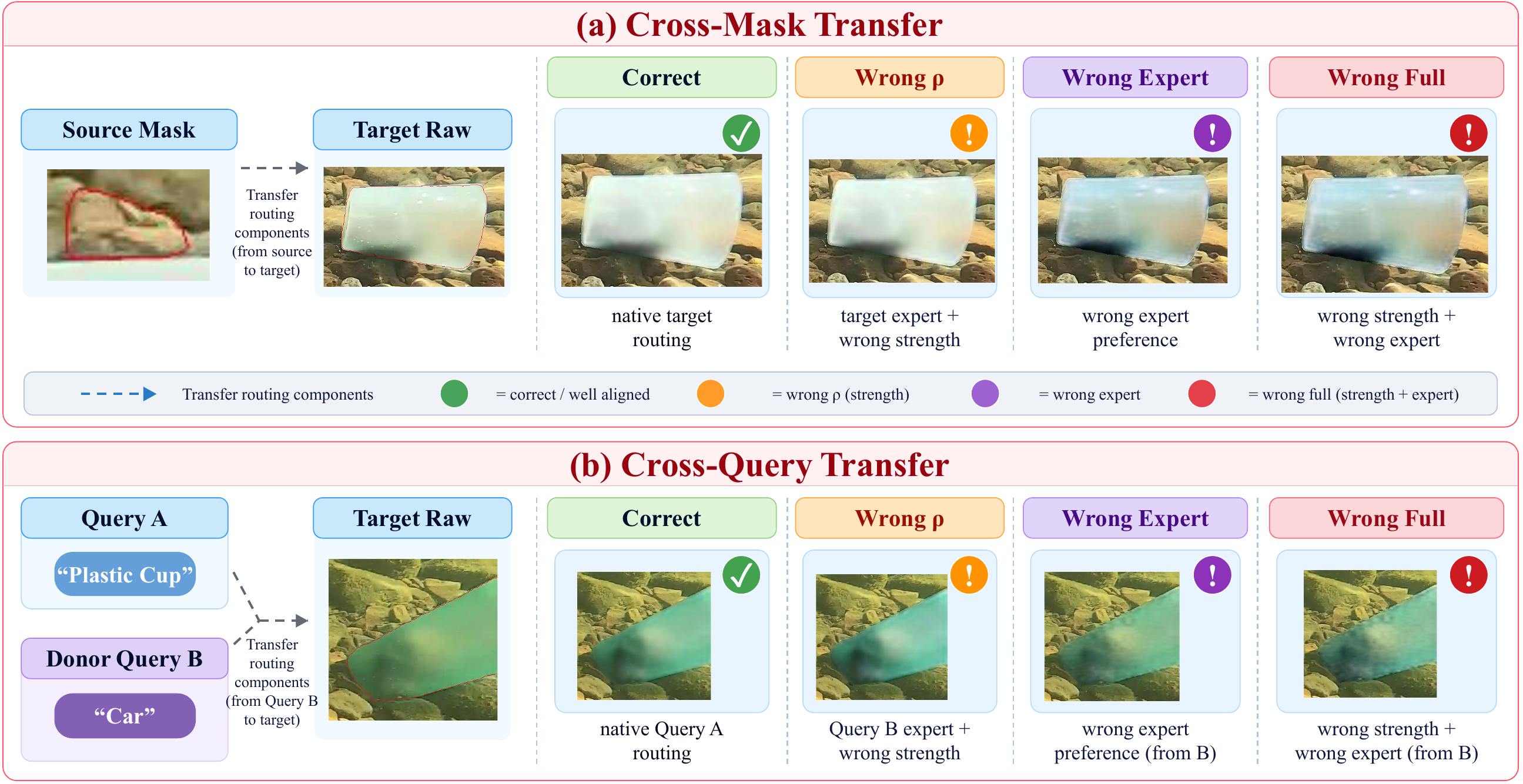}
\caption{Representative cross-mask and cross-query routing-transfer
examples. Target Raw and expert features remain unchanged, while routing
parameters are transferred from another mask or query. Incorrect expert
preference causes larger ranking degradation than transferring enhancement
strength alone.}
\label{fig:routing_transfer_examples}
\end{figure*}

These results show that the preferred enhancement behavior varies across
both candidate masks and language queries. Under the tested transfer settings, transferring expert preference causes a larger mean rank increase than transferring enhancement strength. Representative examples are shown in
Fig.~\ref{fig:routing_transfer_examples}.

\begin{figure}[!t]
\centering
\includegraphics[width=\linewidth]{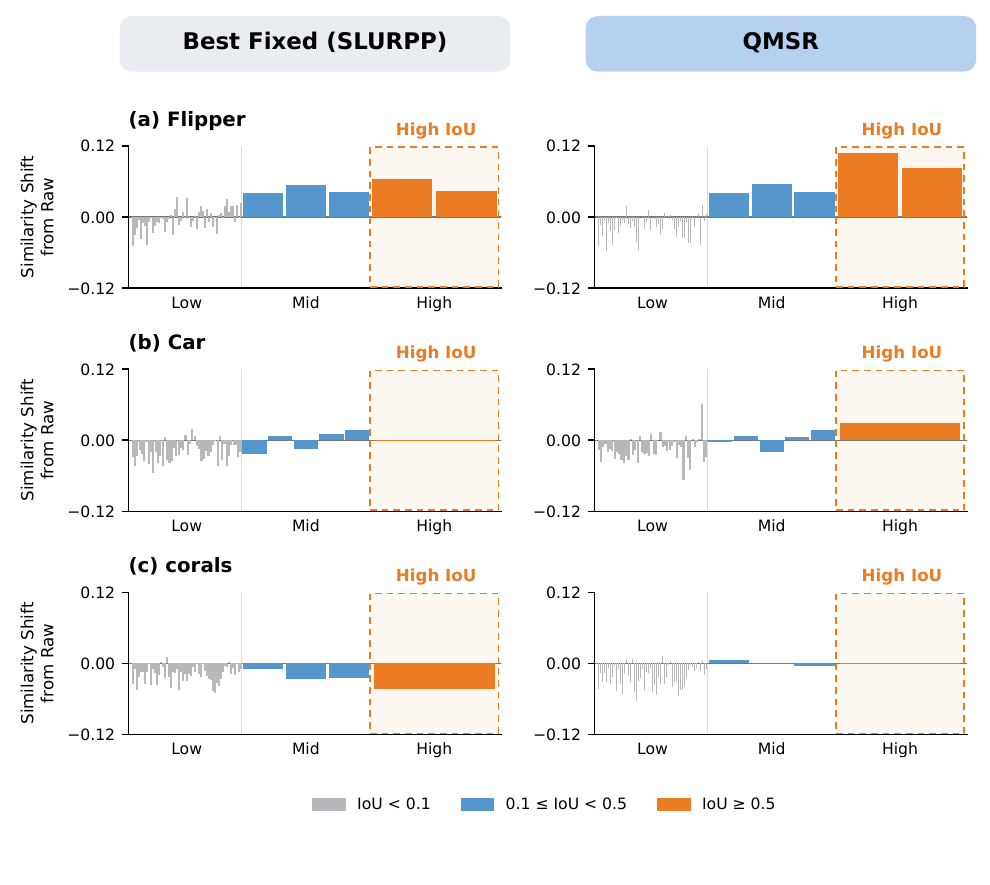}
\caption{Changes in text–image cosine similarity relative to Raw for selected examples, with candidates ordered by IoU using nonuniform spacing. Dashed boxes highlight high-IoU candidates; horizontal zero lines indicate no change from Raw.}
\label{fig:semantic_shift}
\end{figure}

Because these experiments perturb only the routing decision while keeping
the target visual features fixed, they isolate the effect of routing
mismatch from changes in candidate representation.
Fig.~\ref{fig:semantic_shift} illustrates changes in text–image similarity across candidates with different IoU relevance.

\subsection{Analysis of Routing Action Space}
\label{sec:action_space}

We next investigate whether multiple experts are necessary for each
candidate. To eliminate the influence of student optimization, we compare
two privileged Oracle action spaces. The unrestricted Oracle can combine
Raw with all nine experts, whereas the single-expert Oracle is restricted
to Raw plus one selected expert with a continuous strength.

\begin{table}[!t]
\centering
\caption{Privileged action-space comparison.}
\label{tab:oracle_space}
\small
\setlength{\tabcolsep}{8pt}
\begin{tabular}{lcc}
\toprule
Action Space & NDCG@10 & Gain Coverage \\
\midrule
Raw & 0.46548 & -- \\
Raw + Single Expert & 0.74982 & 98.3\% \\
Raw + 9 Experts & 0.75482 & 100\% \\
\bottomrule
\end{tabular}
\end{table}

As shown in Table~\ref{tab:oracle_space}, the single-expert Oracle reaches
0.7498 NDCG@10, compared with 0.7548 for the unrestricted Oracle.
Relative to Raw, the single-expert action space preserves
\begin{equation}
\frac{0.74982-0.46548}
{0.75482-0.46548}
=98.3\%
\end{equation}
of the unrestricted Oracle gain. This result shows that selecting one
query--mask-specific expert together with a continuous Raw--Expert strength
already captures most of the privileged retrieval potential of the current
expert pool.

This does not mean that expert interactions are absent. Under a strict
complementary-rescue criterion, 589 out of 6,141 harmful-alone expert
events become beneficial when combined with other experts, corresponding
to 9.59\%. Such cases occur for all nine experts. Progressively
removing experts from unrestricted compositions results in increasing
performance degradation. These observations indicate that expert
complementarity exists, but its overall contribution is limited for the
current heterogeneous holistic UIE experts.

\subsection{Ablation Study}
\label{sec:ablation}

We investigate the influence of the privileged teacher and the annealed soft-routing relaxation. All student variants in Table~\ref{tab:training_ablation} deploy the same hard Top-1 routing policy.

\begin{table}[!t]
\centering
\caption{Ablation study of the training strategy.}
\label{tab:training_ablation}
\small
\setlength{\tabcolsep}{5pt}
\begin{tabular}{lcc}
\toprule
Teacher / Training & Soft Rank & NDCG@10 \\
\midrule
Single-Expert Teacher & No & 0.57253 \\
Unrestricted Soft Teacher & No & 0.57336 \\
\textbf{Unrestricted Soft Teacher} & \textbf{Yes} & \textbf{0.57857} \\
\bottomrule
\end{tabular}
\end{table}

Using the unrestricted soft teacher instead of a hard single-expert teacher
increases NDCG@10 only slightly from 0.5725 to 0.5734. In contrast,
introducing the annealed soft-routing relaxation further improves the
hard Top-1 student to 0.5786. Therefore, the main benefit does not come
from deploying a soft expert mixture, but from providing a smoother
optimization path for the discrete expert-selection problem.

The final model reaches 0.5828 validation NDCG@10 and stops at epoch 86.
After validation-only Raw-fallback calibration, $\tau_\rho=0.23$ is fixed
for test evaluation. The resulting router uses different experts across
the dataset without collapsing to a single dominant expert.

\section{Conclusion}

We proposed QMSR, a query-conditioned mask-wise expert routing framework
for open-vocabulary underwater object retrieval. For each candidate, QMSR
predicts an enhancement expert and a continuous Raw--Expert strength from
the language query, Raw mask feature, and candidate-set context.
Raw-anchored fusion exploits useful enhancement while preserving the Raw
representation when enhancement is unnecessary or harmful. The discrete
router is trained with privileged Oracle supervision and annealed soft
routing, while deployed routing uses neither enhanced features nor
ground-truth IoU. Experiments show that QMSR improves NDCG@10 by 26.4\%
over the best fixed UIE expert and by 17.6\% over the image--query shared
router. On category-disjoint unseen queries, QMSR further achieves an
8.98\% improvement over the best fixed UIE expert. Oracle analysis shows
that the single-expert action space preserves 98.3\% of the gain achieved
by unrestricted multi-expert fusion. These results demonstrate the
effectiveness of query-conditioned, candidate-specific, and Raw-anchored
enhancement routing for robust underwater open-vocabulary retrieval.
\section*{ACKNOWLEDGMENTS}

\textbf{AI Usage Statement.} OpenAI Codex assisted with experimental code
development, experiment execution, and manuscript drafting and editing.
The authors take full responsibility for the reported results and final
manuscript.
\balance
\bibliographystyle{IEEEtran}
\bibliography{qmsr_refs}

@inproceedings{radford2021clip,
  title={Learning Transferable Visual Models From Natural Language Supervision},
  author={Radford, Alec and Kim, Jong Wook and Hallacy, Chris and others},
  booktitle={Proc. ICML},
  year={2021}
}

@inproceedings{kirillov2023sam,
  title={Segment Anything},
  author={Kirillov, Alexander and Mintun, Eric and Ravi, Nikhila and others},
  booktitle={Proc. ICCV},
  year={2023}
}

@inproceedings{jatavallabhula2023conceptfusion,
  title={{ConceptFusion}: Open-Set Multimodal {3D} Mapping},
  author={Jatavallabhula, Krishna Murthy and Kuwajerwala, Alihusein and Gu, Qiao and others},
  booktitle={Proc. RSS},
  year={2023}
}

@article{wen2026plaf,
  title={{PLAF}: Pixel-wise Language-Aligned Feature Extraction for Efficient {3D} Scene Understanding},
  author={Wen, Junjie and He, Junlin and Ma, Fei and Cui, Jinqiang},
  journal={arXiv preprint arXiv:2604.15770},
  year={2026}
}

@inproceedings{akkaynak2019seathru,
  title={Sea-thru: A Method for Removing Water From Underwater Images},
  author={Akkaynak, Derya and Treibitz, Tali},
  booktitle={Proc. CVPR},
  year={2019}
}

@article{islam2020funie,
  title={Fast Underwater Image Enhancement for Improved Visual Perception},
  author={Islam, Md Jahidul and Xia, Youya and Sattar, Junaed},
  journal={IEEE Robot. Autom. Lett.},
  year={2020}
}

@article{li2021ucolor,
  title={Underwater Image Enhancement via Medium Transmission-Guided Multi-Color Space Embedding},
  author={Li, Chongyi and others},
  journal={IEEE Trans. Image Process.},
  year={2021}
}

@article{peng2023ushape,
  title={U-Shape Transformer for Underwater Image Enhancement},
  author={Peng, Lintao and Zhu, Chunli and Bian, Liheng},
  journal={IEEE Trans. Image Process.},
  year={2023}
}

@inproceedings{huang2023semiuir,
  title={Contrastive Semi-Supervised Learning for Underwater Image Restoration via Reliable Bank},
  author={Huang, Shirui and others},
  booktitle={Proc. CVPR},
  year={2023}
}

@article{cong2023pugan,
  author={Cong, Runmin and others},
  title={PUGAN: Physical Model-Guided Underwater Image Enhancement Using GAN With Dual-Discriminators},
  journal={IEEE Trans. Image Process.},
  year={2023}
}

@inproceedings{zhao2024wfdiff,
  title={Wavelet-based Fourier Information Interaction with Frequency Diffusion Adjustment for Underwater Image Restoration},
  author={Zhao, Chen and others},
  booktitle={Proc. CVPR},
  year={2024}
}

@article{guan2024watermamba,
  title={WaterMamba: Visual State Space Model for Underwater Image Enhancement},
  author={Guan, Meisheng and others},
  journal={arXiv preprint arXiv:2405.08419},
  year={2024}
}

@inproceedings{peng2025ssuie,
  title={Adaptive Dual-domain Learning for Underwater Image Enhancement},
  author={Peng, Lintao and Bian, Liheng},
  booktitle={Proc. AAAI},
  year={2025}
}

@article{wu2025slurpp,
  title={Single-Step Latent Diffusion for Underwater Image Restoration},
  author={Wu, Jiayi and others},
  journal={IEEE Trans. Pattern Anal. Mach. Intell.},
  year={2025}
}

@article{lin2026dtiuie,
  title={Downstream Task-Inspired Underwater Image Enhancement: A Perception-Aware Study From Dataset Construction to Network Design},
  author={Lin, Bosen and others},
  journal={IEEE Trans. Image Process.},
  year={2026}
}

@inproceedings{fan2026semantic,
  title={Empowering Semantic-Sensitive Underwater Image Enhancement with VLM},
  author={Fan, Guodong and others},
  booktitle={Proc. AAAI},
  year={2026}
}

@article{hong2020trashcan,
  title={TrashCan: A Semantically-Segmented Dataset towards Visual Detection of Marine Debris},
  author={Hong, Jungseok and Fulton, Michael and Sattar, Junaed},
  journal={arXiv preprint arXiv:2007.08097},
  year={2020}
}

@inproceedings{lian2024usis,
  title={Diving into Underwater: Segment Anything Model Guided Underwater Salient Instance Segmentation and A Large-scale Dataset},
  author={Lian, Shijie and others},
  booktitle={Proc. ICML},
  year={2024}
}

@article{li2025uiis10k,
  title={Advancing Marine Research: UWSAM Framework and UIIS10K Dataset for Precise Underwater Instance Segmentation},
  author={Li, Hua and others},
  journal={arXiv preprint arXiv:2505.15581},
  year={2025}
}

@inproceedings{mukherjee2025cou,
  title={The Common Objects Underwater (COU) Dataset for Robust Underwater Object Detection},
  author={Mukherjee, Rishi and others},
  booktitle={Proc. IROS},
  year={2025}
}

@inproceedings{li2026maris,
  title={MARIS: Marine Open-Vocabulary Instance Segmentation},
  author={Li, Bingyu and Wang, Feiyu and Zhang, Da and Zhao, Zhiyuan and Gao, Junyu and Li, Xuelong},
  booktitle={Proc. CVPR},
  year={2026}
}

@inproceedings{li2026earth2ocean,
  title={Exploring the Underwater World Segmentation without Extra Training},
  author={Li, Bingyu and Huo, Tao and Zhang, Da and Zhao, Zhiyuan and Gao, Junyu and Li, Xuelong},
  booktitle={Proc. CVPR},
  year={2026}
}

@inproceedings{shazeer2017moe,
  title={Outrageously Large Neural Networks: The Sparsely-Gated Mixture-of-Experts Layer},
  author={Shazeer, Noam and others},
  booktitle={Proc. ICLR},
  year={2017}
}

@article{fedus2022switch,
  title={Switch Transformers: Scaling to Trillion Parameter Models with Simple and Efficient Sparsity},
  author={Fedus, William and Zoph, Barret and Shazeer, Noam},
  journal={J. Mach. Learn. Res.},
  year={2022}
}

@inproceedings{riquelme2021vmoe,
  title={Scaling Vision with Sparse Mixture of Experts},
  author={Riquelme, Carlos and others},
  booktitle={Proc. NeurIPS},
  year={2021}
}

@inproceedings{mustafa2022limoe,
  title={Multimodal Contrastive Learning with LIMoE: The Language-Image Mixture of Experts},
  author={Mustafa, Basil and others},
  booktitle={Proc. NeurIPS},
  year={2022}
}

@inproceedings{lee2025routerretriever,
  title={RouterRetriever: Routing over a Mixture of Expert Embedding Models},
  author={Lee, Hyunji and others},
  booktitle={Proc. AAAI},
  year={2025}
}

@article{abdallah2026argus,
  author={Abdallah, Abdelrahman and others},
  title={Argus-Retriever: Vision-LLM Late-Interaction Retrieval with Region-Aware Query-Conditioned MoE for Visual Document Retrieval},
  journal={arXiv preprint arXiv:2606.04300},
  year={2026}
}

@inproceedings{li2022airnet,
  title={All-in-One Image Restoration for Unknown Corruption},
  author={Li, Boyun and others},
  booktitle={Proc. CVPR},
  year={2022}
}

@inproceedings{potlapalli2023promptir,
  title={PromptIR: Prompting for All-in-One Blind Image Restoration},
  author={Potlapalli, Vaishnav and others},
  booktitle={Proc. NeurIPS},
  year={2023}
}

@inproceedings{kim2020spatialmoe,
  title={Restoring Spatially-Heterogeneous Distortions Using Mixture of Experts Network},
  author={Kim, Sijin and Ahn, Namhyuk and Sohn, Kyung-Ah},
  booktitle={Proc. ACCV},
  year={2020}
}

@article{zhang2025uniuir,
  title={UniUIR: Considering Underwater Image Restoration as an All-in-One Learner},
  author={Zhang, Xu and others},
  journal={IEEE Trans. Image Process.},
  year={2025}
}

@article{wang2025m2restore,
  title={M2Restore: Mixture-of-Experts-Based Mamba-CNN Fusion Framework for All-in-One Image Restoration},
  author={Wang, Yongzhen and others},
  journal={IEEE Trans. Image Process.},
  year={2025}
}

@article{kong2026coreuie,
  title={CoRe-UIE: Rethinking Coexisting and Region-wise Degradation for Underwater Image Enhancement},
  author={Kong, Weifeng and others},
  journal={arXiv preprint arXiv:2608.08965},
  year={2026}
}

@inproceedings{liu2022iayolo,
  title={Image-Adaptive YOLO for Object Detection in Adverse Weather Conditions},
  author={Liu, Wenyu and others},
  booktitle={Proc. AAAI},
  year={2022}
}

@inproceedings{wen2024enyolo,
  title={EnYOLO: A Real-Time Framework for Domain-Adaptive Underwater Object Detection with Image Enhancement},
  author={Wen, Junjie and others},
  booktitle={Proc. ICRA},
  year={2024}
}

@article{huang2026amieod,
  title={AMIEOD: Adaptive Multi-Experts Image Enhancement for Object Detection in Low-Illumination Scenes},
  author={Huang, Xiaochen and others},
  journal={arXiv preprint arXiv:2605.06084},
  year={2026}
}

@inproceedings{conde2024instructir,
  title={InstructIR: High-Quality Image Restoration Following Human Instructions},
  author={Conde, Marcos V. and Geigle, Gregor and Timofte, Radu},
  booktitle={Proc. ECCV},
  year={2024}
}

@article{liu2025textguided,
  title={Text-guided Image Restoration and Semantic Enhancement for Text-to-Image Person Retrieval},
  author={Liu, Delong and others},
  journal={Neural Networks},
  year={2025}
}

@article{phung2024shared,
  author={Phung, Amy and others},
  title={A Shared Autonomy System for Precise and Efficient Remote Underwater Manipulation},
  journal={IEEE Trans. Robot.},
  volume={40},
  pages={4147--4159},
  year={2024}
}

@article{wang2022underwater,
  author={Wang, Haihang and others},
  title={A Bidirectional Soft Biomimetic Hand Driven by Water Hydraulic for Dexterous Underwater Grasping},
  journal={IEEE Robot. Autom. Lett.},
  year={2022}
}

@inproceedings{singh2025openset,
  title={Open-Set Semantic Uncertainty Aware Metric-Semantic Graph Matching},
  author={Singh, Kurran and Leonard, John J.},
  booktitle={Proc. ICRA},
  year={2025}
}

@inproceedings{thengane2025merlion,
  title={{MERLION}: Marine ExploRation with Language guIded Online iNformative Visual Sampling and Enhancement},
  author={Thengane, Shrutika Vishal and Prasetyo, Marcel Bartholomeus and Tan, Yu Xiang and Meghjani, Malika},
  booktitle={Proc. ICRA},
  year={2025}
}

@article{qi2026airembodied,
  title={AIR-Embodied: Active Interactive Reconstruction for 3D Gaussian Splatting with Embodied Multimodal Agents},
  author={Qi, Zhenghao and others},
  journal={Unmanned Systems},
  year={2026}
}

\end{document}